\documentclass[runningheads]{llncs}

\usepackage{eccv}

\usepackage{eccvabbrv}

\usepackage{graphicx}
\usepackage{booktabs}
\usepackage{tabularx} 
\usepackage{booktabs} 
\usepackage{multirow}
\usepackage{array}
\usepackage[accsupp]{axessibility}  

\usepackage{color}
\definecolor{hs}{rgb}{0,0,0}

\usepackage[pagebackref,breaklinks,colorlinks,citecolor=eccvblue]{hyperref}

\usepackage{orcidlink}

\begin{document}

\title{Emerging Property of Masked Token \\for Effective Pre-training}

\titlerunning{MTO}

\author{Hyesong Choi\inst{1} \and Hunsang Lee\inst{2} \and Seyoung Joung\inst{1} \and \\
Hyejin Park\inst{1} \and Jiyeong Kim\inst{1} \and Dongbo Min\inst{1}}

\authorrunning{H.~Choi \textit{et al}.}

\institute{Ewha W. University \and Hyundai Motor Company}

\maketitle

\begin{abstract}
Driven by the success of Masked Language Modeling (MLM), the realm of self-supervised learning for computer vision has been invigorated by the central role of Masked Image Modeling (MIM) in driving recent breakthroughs. Notwithstanding the achievements of MIM across various downstream tasks, its overall efficiency is occasionally hampered by the lengthy duration of the pre-training phase. This paper presents a perspective that the optimization of masked tokens as a means of addressing the prevailing issue. Initially, we delve into an exploration of the inherent properties that a masked token ought to possess. Within the properties, we principally dedicated to articulating and emphasizing the `data singularity' attribute inherent in masked tokens. Through a comprehensive analysis of the heterogeneity between masked tokens and visible tokens within pre-trained models, we propose a novel approach termed \textbf{masked token optimization (MTO)}, specifically designed to improve model efficiency through weight recalibration and the enhancement of the key property of masked tokens. The proposed method serves as an adaptable solution that seamlessly integrates into any MIM approach that leverages masked tokens. As a result, MTO achieves a considerable improvement in pre-training efficiency, resulting in an approximately 50\% reduction in pre-training epochs required to attain converged performance of the recent approaches. 
\end{abstract}

\section{Introduction}~\label{sec:intro}
Pre-training of universal language representations~\cite{radford2018improving, devlin2018bert, liu2019roberta, clark2020electra, raffel2020exploring} has been a crucial area of Natural Language Processing (NLP), especially when training large-scale models~\cite{brown2020language, taylor2022galactica}. Following the philosophy of Masked Language Modeling (MLM)~\cite{devlin2018bert, liu2019roberta, clark2020electra, bao2020unilmv2}, Masked Image Modeling (MIM)~\cite{bao2021beit, he2022masked, xie2022simmim, dong2022bootstrapped} has been at the core of recent advances in self-supervised learning for computer vision. MIM applies the principles of MLM to images, enabling effective pre-training of Transformers and improving transfer learning performances. The essence of these Masked Signal Modeling approaches lies in encouraging the model to predict the gaps in an input signal to learn the contextual relationships between signals while capturing an overall structure. 

Despite the tremendous successes of MIM in diverse downstream tasks, the long pre-training phase that it entails tends to impede its efficiency. Concretely, a substantial amount of pre-training, typically from 800 to 1600 epochs, is essential to attain the convergence of the Transformer for transfer learning. Meanwhile, several methodologies have been employed for MIM in an effort to alleviate the disparities that exist between the linguistic and visual domains when leveraging the MLM concept. For instance, patch tokenization~\cite{bao2021beit} is introduced to emulate the discrete nature of language tokens, while raw pixel regression~\cite{xie2022simmim} is adopted to attune with the continuous visual signals. However, the intrinsic properties of masked tokens, a vital component of MIM, have yet to be comprehensively surveyed by the vision community. 
In this paper, we cast the lengthy pre-training issue due to the low convergence rate from the perspective of the optimization of masked tokens which arises as 
lack of consideration of the inherent properties of masked tokens.


Here we come to the pivotal inquiry that lies at the heart of the discourse: \textit{What properties should a masked token have within the realm of MIM?} Given the premise that the masked token is selected and masked from the training data, it is imperative that the selected masked token exhibits certain specific attributes; (i) Spatial Randomness: Masked tokens must be randomly selected from the corpus of input patches, so that the model can learn to predict tokens in various locations and semantics. Regarding this property, a research direction incorporating prior knowledge into the spatial randomness of masked tokens~\cite{kakogeorgiou2022hide, wu2022object} is currently the subject of \textcolor{hs}{animated debate}. (ii) Substitutional Consistency, \textcolor{hs}{in the masking process, randomly selected visible tokens} should consistently be replaced with the same learnable parameters~\cite{wettig2023should}. This allows the model to easily recognize the masked tokens and learn to reconstruct them during pre-training. Lastly, (iii) \textbf{Data Singularity}. This last facet represents a novel property that we aim to assert and \textcolor{hs}{demonstrate} throughout the entire manuscript. It signifies that the \textcolor{hs}{masked token} in the initial embedding should be unique \textcolor{hs}{token} that are unlikely to manifest in the training data. Stated differently, the masked tokens should exhibit a \textcolor{hs}{negligible} correlation with visible tokens to mitigate the possibility of obfuscation, when given as inputs to the attention layers. Employing masked tokens that are well differentiated from visible tokens enables the model to identify semantics within the training data, thereby improving focused pretext prediction capability.

Visual signals are inherently continuous, making it challenging for masked tokens to ensure data singularity, as they cannot be explicitly differentiated like their discrete text token counterparts. To be specific, due to the clear semantics associated with each word in the text, the distinctiveness of masked tokens can be easily preserved during the pretext prediction process in the linguistic domain. Contrarily, the Tokenizer-based approach~\cite{li2022mc} has reported that in image tokenizers, different semantic patches can have similar token under visual discretization. This finding indicates that the task of distinguishing masked tokens from visible tokens is \textcolor{hs}{adverse} in the context of the visual tokenizer, where patches are represented as continuous values. Hence, attaining the desired distinctiveness of the masked token to the training data solely relies upon the model's convergence through a \textcolor{hs}{prolonged} pre-training process, akin to the predictions of a black-box system devoid of explicit constraints. Therefore, we propose an analysis of masked tokens and optimization based on it by directing our attention toward the data singularity among the trifecta of properties. 


Our initial step encompassed a heterogeneity analysis of masked tokens against the visible tokens to demonstrate the manifestation of the masked token's data singularity characteristic within the model upon reaching convergence. Moreover, the scope of this analysis is designed to investigate both the extent and the tendency of how heterogeneity unfolds throughout the different layers of the network's architecture.
Building upon the insights from the heterogeneity analysis, we propose a sophisticated method for optimizing masked tokens. The proposed Masked Token Optimization (MTO) approach includes a strategic exclusion of semantically inconsequential masked tokens from the weight aggregation process associated with visible tokens, achieved through weight recalibration. At the same time, the proposed MTO method explicitly imposes constraints on data singularity throughout the optimization of masked tokens to reinforce the model's capacity to differentiate between tasks, given the distinctive roles that masked tokens and visible tokens assume within the architecture; the masked token is integral to pretext prediction, whereas the visible token is essential for the encoding and decoding of representations.


The proposed Masked Token Optimization (MTO) represents a versatile and adaptable method capable of seamless integration into any MIM-based approach utilizing masked tokens, thus empowering pre-training operations with heightened efficiency and performance. The succeeding sections of the paper \textcolor{hs}{present} the empirical evidence of the efficacy of the MTO approach when integrated into diverse MIM methodologies including SimMIM~\cite{xie2022simmim}, MAE~\cite{he2022masked} and BootMAE~\cite{dong2022bootstrapped}. The findings demonstrate that the application of MTO induces rapid model convergence and substantial improvements in representation learning. Notably, MTO improves the pre-training efficiency by approximately halving the pre-training epochs required to reach converged performance in the recent MIM approaches. Such outcomes provide a compelling justification for the wide-scale adoption of MTO as an useful plug-and-play tool in pre-training procedures.

\section{Preliminaries}
We start by revisiting the recent framework of MIM. The latest advancements in MIM~\cite{he2022masked, xie2022simmim, dong2022bootstrapped} have surpassed the past two-stage methods~\cite{bao2021beit, li2022mc} by integrating masked prediction and the autoencoder training in a single end-to-end process, aiming at encoding valuable representation and predicting pretext for masked patches~\cite{zhang2022survey}. As these approaches are built upon Transformer architectures~\cite{vaswani2017attention, dosovitskiy2020image, liu2021swin, liu2022swin}, we assume the underlying framework is an attention model~\cite{dosovitskiy2020image, liu2021swin} throughout the paper. An input image $\textit{I} \in \mathbb{R}^{HW \times 3}$ is first divided into non-overlapping $N=H \times W / P^{2}$ patches. Then, patches are randomly sampled and masked with a high masking ratio, reflecting the information redundancy~\cite{he2022masked}.

Let $\delta_{M}$ be defined as a set of masked indexes where visible tokens are replaced by a mask token. 
In general, the representation of masked modeling is trained via the minimization of the following self-supervised objective: 

\begin{equation}
    \mathcal{L}_{ss}(f(I;\Theta))=\frac{1}{|\delta_{M}|} \sum_{i \in \delta_{M}} \| f(I;\Theta)_{i} - I_{i} \|^{2}_{2}, 
\end{equation}

\noindent with pretext prediction network $f$ and its learnable parameters $\Theta$. In SimMIM~\cite{xie2022simmim}, $f$ is jointly learned with semantic encoding using masked tokens within the encoder. In contrast, the encoder of MAE~\cite{he2022masked} solely leverages the visible image tokens. Then, the encoded visual patches are fed into the Transformer decoder, where masked tokens are employed for pretext prediction.



\section{Analysis}\label{sec:analysis}
To investigate the tendency of the heterogeneity between masked token and visible tokens, we analyze the pre-trained models of the recent approaches~\cite{he2022masked, xie2022simmim}.

\subsection{Heterogeneity Measure via Entropy}
We define the degree of heterogeneity as the mutual dependence of masked tokens with respect to visible tokens in each layer as follows:

\begin{equation}
\begin{aligned}
    H = -\frac{1}{|\delta_{M}|} \sum_{i} \sum_{j} A_{i,j} \text{log} (A_{i,j}) \quad\quad \text{where} ~~ A = \psi(X_{M} X_{V}^{\top}).
    \label{eq:heterogeneity_measure}
\end{aligned}
\end{equation}
We define $X_{M}=\{x_{i} | i \in \delta_{M}\}$ as a set of masked tokens, $X_{V}=\{x_{i} | i \notin \delta_{M}\}$ as a set of visible image token, 
$A \in \mathbb{R}^{|\delta_{M}| \times (N - |\delta_{M}|)}$ is the affinity matrix that represents the probabilistic similarity between $X_{M}$ and $X_{V}$, and $\psi$ is the scaling function, \emph{i.e.} row-wise softmax, that scales logits into a probability distribution relative to the visible image token. 

\subsection{Heterogeneity Analysis}
We assume that it is necessary for masked tokens and visible tokens to exhibit substantial variability in terms of their distinct data properties in the initial embedding prior to being processed by the attention layers. This distinctiveness between masked tokens and visible tokens is instrumental in enhancing the model's ability to differentiate between the two tasks, owing to their distinct roles in the architecture; the former serves the purpose of pretext prediction while the latter aids in feature encoding and decoding. On the other hand, in subsequent layers, the masked tokens are gradually recovered by the neighboring visible tokens and will exhibit a heightened correlation with them.

Our investigation centered on determining whether models that demonstrate effective convergence uphold these hypotheses. To this end, we analyze the pre-trained models of the recent approaches~\cite{he2022masked, xie2022simmim} from two facets: 1) Heterogeneity analysis on pre-trained models of different methods and 2) heterogeneity analysis of converged and non-converged models.

\begin{figure*}[t!]
	\centering
	\includegraphics[width=0.8\columnwidth]{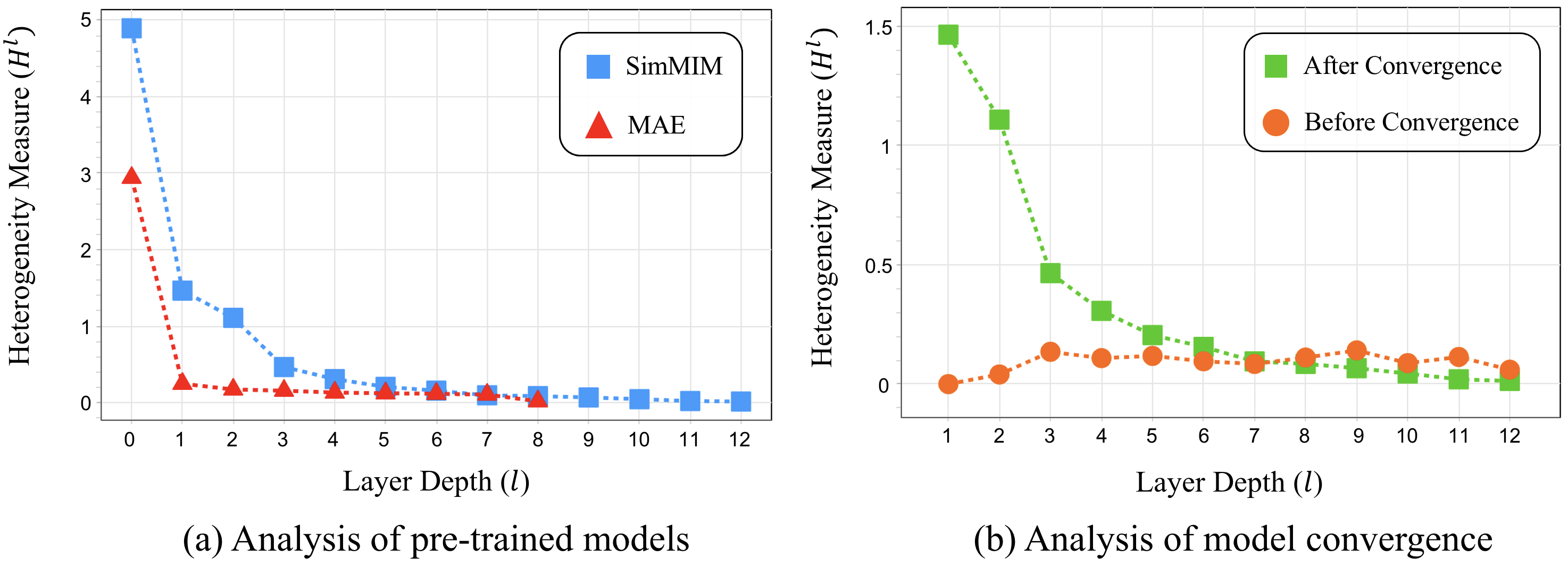}
	\caption{To investigate the heterogeneity between masked token and visible token, we analyze the pre-trained models of the recent approaches~\cite{xie2022simmim, he2022masked}. (a) shows that the heterogeneity between two distinct types of tokens is highest on the initial embedding for both approaches, and it gradually decreases in subsequent layers. Unlike the pre-trained model, the heterogeneity of the non-converged model shown in (b) displays an erratic trend, indicating that the tendency of heterogeneity is acquired through model convergence.
	}
	\label{fig:heterogeneity_pretrained}
	\vspace{-0.4cm}
\end{figure*}

\begin{table*}[t]
    \centering
    \addtolength{\tabcolsep}{-2pt}
    \renewcommand{\arraystretch}{0.8}
    \begin{tabular}{lccccccccccccc}
        \toprule
        Layer Depth   &   0 &   1 &   2 &   3 &  4 & 5 & 6 & 7 & 8 & 9 & 10 & 11 & 12\\ 
        \midrule
        SimMIM~\cite{xie2022simmim}  &   ~4.90~ & ~1.46~ & ~1.11~ & ~0.46~ & ~0.31~ & ~0.21~ & ~0.16~ & ~0.10~ & ~0.08~ & ~0.07~ & ~0.04~ & ~0.02~ & ~0.01~ \\
        \midrule
        MAE~\cite{he2022masked}    & 2.94 & 0.25 & 0.17 &   0.16 & 0.13 & 0.12 & 0.11 & 0.10 & 0.02 & - & - & - & -\\
        \midrule
    \end{tabular}
    \vspace{0.1cm}
    \caption{We present the specific values for the heterogeneity measure of Figure~\ref{fig:heterogeneity_pretrained} (a). The heterogeneity from the initial input to the subsequent layers shows a distinct decrease in both methods~\cite{xie2022simmim, he2022masked}. 
    }
    \label{tab:hetereo_pretrained}
    \vspace{-0.8cm}
\end{table*}

\subsubsection{Heterogeneity analysis on pre-trained models of different methods}
The heterogeneity between the masked and visible tokens across every layer of two pre-trained models, MAE and SimMIM~\cite{he2022masked, xie2022simmim} utilizing ViT~\cite{dosovitskiy2020image} as a backbone is shown in Figure~\ref{fig:heterogeneity_pretrained} (a). Note that `layer depth 0' refers to the initial embedding stage before the masked token is fed into the attention layer as an input. Figure~\ref{fig:heterogeneity_pretrained} (a) shows that the heterogeneity between two distinct types of tokens is highest on the initial embedding for both approaches, and it gradually decreases in subsequent layers. Specific values for the heterogeneity measure of each layer in Figure~\ref{fig:heterogeneity_pretrained} (a) can be found in Table~\ref{tab:hetereo_pretrained}. The high heterogeneity in the initial embedding phase reflects the data singularity of the masked tokens, whereas the masked tokens are reconstructed to resemble the visible tokens, leading to reduced heterogeneity in subsequent layers. 

\begin{figure*}[t!]
	\centering
	\includegraphics[width=0.9\columnwidth]{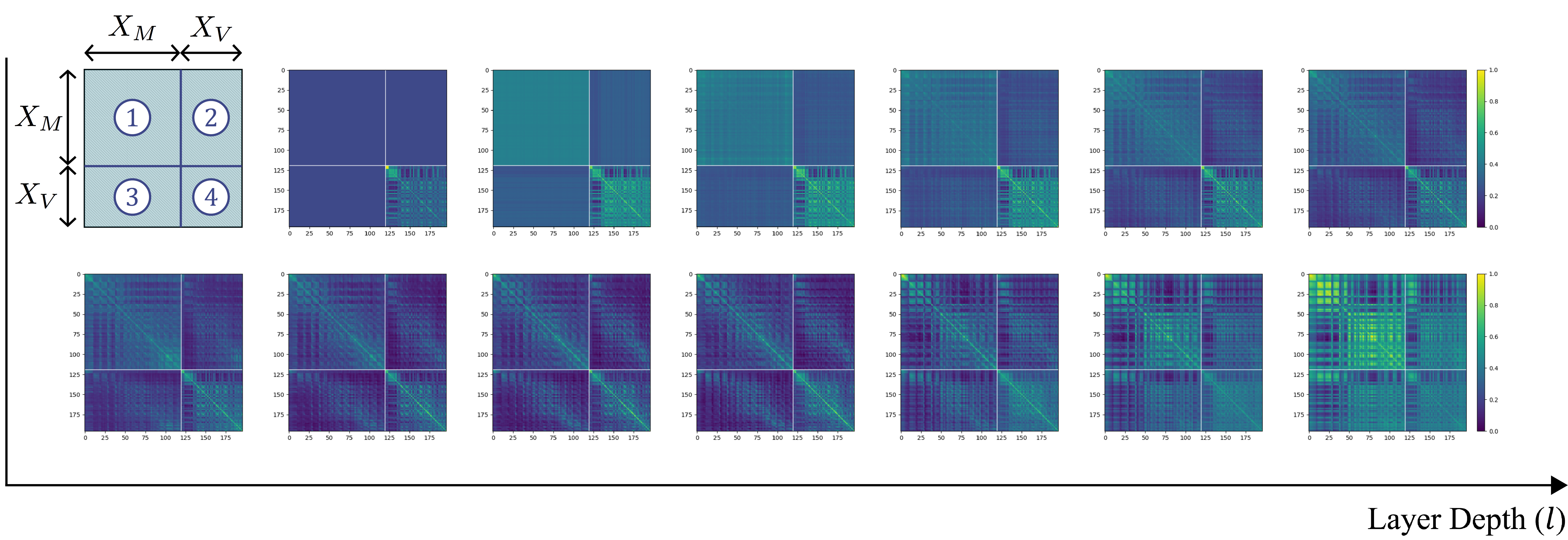}
	\caption{We present the affinity map between every token pair for each layer of the pre-trained model~\cite{xie2022simmim}. Affinity maps are listed in order from initial embedding to subsequent layers, and the x-axis and y-axis of the affinity map are both arranged in the order of masked token $X_M$ and the visible image token $X_V$. Min-max normalization was used for the visualization of the affinity maps. 
	}
	\label{fig:affinty_matrix}
	\vspace{-0.4cm}
\end{figure*}

Furthermore, Figure~\ref{fig:affinty_matrix} presents the affinity map between every token pair 
for each layer of the pre-trained model~\cite{xie2022simmim}. Affinity maps are listed in order from initial embedding to subsequent layers, and we used min-max normalization for visualization. In the interest of understanding, the x-axis and y-axis of the affinity map are both arranged in the order of masked token $X_M$ and visible image token $X_V$. Thus, we divided the affinity map into quadrants. Note that, the heterogeneity $H$ is defined in the second quadrant as it corresponds to the mutual dependence of masked tokens with respect to visible tokens.

In the first quadrant, the masked token, initially a singular parameter in the initial embedding, resembles the visible token in the subsequent layers, resulting in various correlation values. Within the second and third quadrants, the relationship between the masked token and the visible token exhibits a relatively low correlation during the initial layers. However, as the masked token undergoes the reconstruction process, increasingly higher correlations materialize in the subsequent layers.
Finally, the correlation between the visible tokens in the fourth quadrant is relatively constant regardless of layer depth. In conclusion, the result of the affinity map qualitatively validates the orthogonality between the masked token and the visible token in the initial embedding while revealing a progressive enhancement in their similarity throughout the subsequent layers. In line with the above hypothesis, the heterogeneity from the initial input to the subsequent layers shows a distinct decrease in the overall results of Figure~\ref{fig:heterogeneity_pretrained}, Table~\ref{tab:hetereo_pretrained}, and Figure~\ref{fig:affinty_matrix}.

\subsubsection{Heterogeneity analysis of converged and non-converged models}
The heterogeneity of the converged model (`After Convergence') and the non-converged model at the early stage (`Before Convergence') of SimMIM~\cite{xie2022simmim} is shown in Figure~\ref{fig:heterogeneity_pretrained} (b). Unlike the converged model, where the heterogeneity steadily decreases, the heterogeneity of the non-converged model displays an erratic trend, lacking any discernible pattern or structure. The results indicate that a model lacking a distinct inclination towards heterogeneity at the beginning of training achieves the desired attributes of the masked token through subsequent convergence.

\section{Masked Token Optimization}
Our endeavor lies in mitigating the issue of the \textcolor{hs}{prolonged} pre-training phase by imposing explicit constraints on the optimization of masked tokens.

In the initial embedding, the parameter designated for masked tokens is identical across all corrupted tokens, 
with their respective values being set via a random initialization process. 
In this context, the \textit{semantic voidness} within masked tokens exerts a detrimental impact on the process of learning features of visible tokens, thereby impeding the overall efficacy of the representation learning. Therefore, we propose an explicit optimization in the initial embedding stage that allows the masked token to be reconstructed by being influenced by the visible token, but conversely, constrains the visible token from being affected by the masked token parameter. This is achievable through the integration of a sparsity-inducing constraining term directly into the weight-learning mechanism of the affinity matrix between masked tokens and visible tokens. As shown in Figure~\ref{fig:affinty_matrix}, the x-axis and y-axis of the affinity map are both arranged in the order of masked token $X_M$ and visible image token $X_V$, which allows the affinity map to be divided into four quadrants. Concretely, we propose to explore intuitive per-row sparsities within the third and fourth quadrants of the matrix as they correspond to the reciprocal dependencies between image tokens in relation to masked tokens and between image tokens themselves, respectively. The following constraint recalibrates the weight distribution between visible and masked tokens on a row-specific basis, increasing the weight between visible-visible tokens in comparison to the weight assigned to visible-masked token interactions:

\begin{figure}[t!]
    \includegraphics[width=0.6\columnwidth]{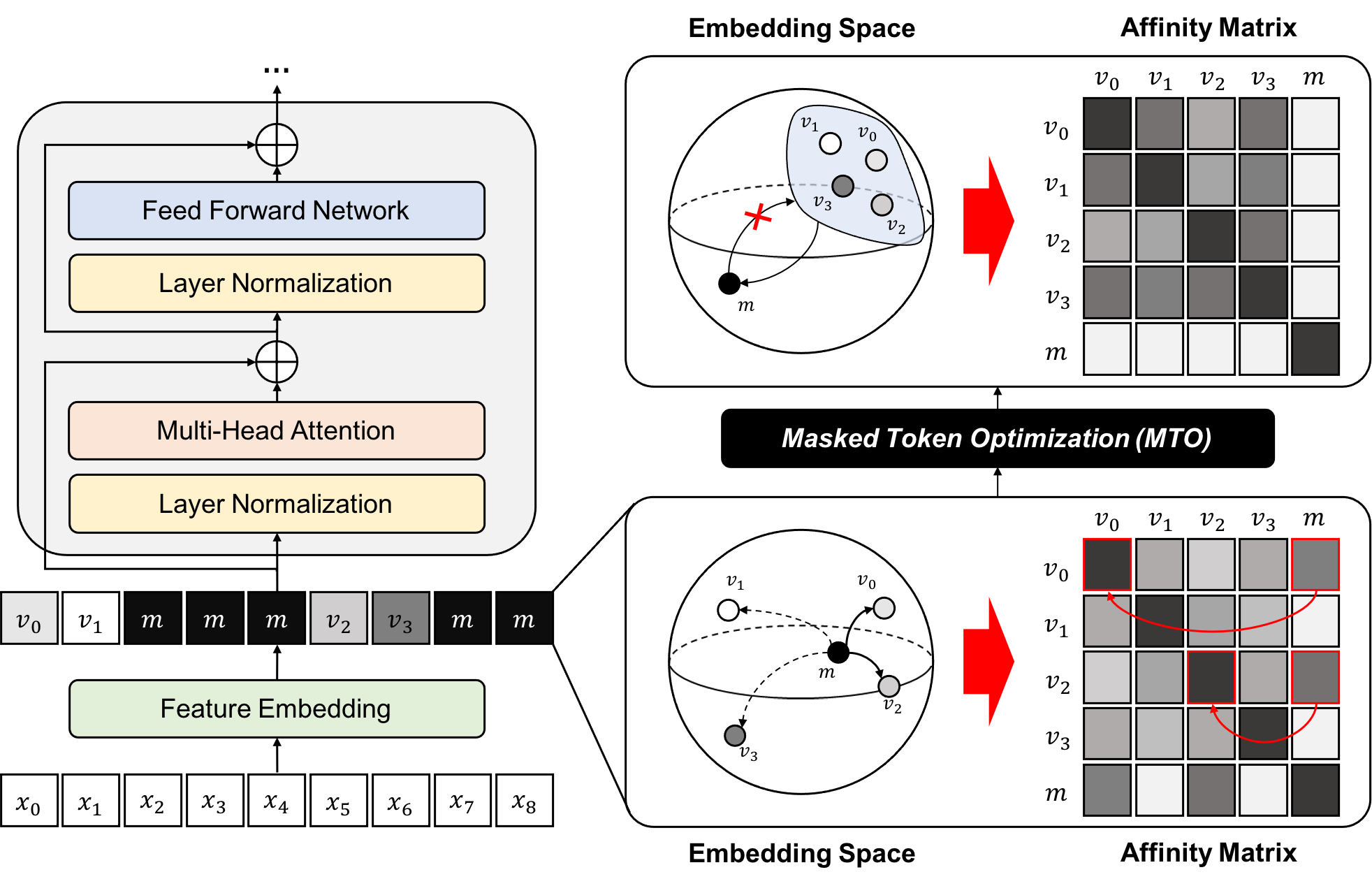}
    \centering
    \caption{The proposed Masked Token Optimization (MTO) approach encompasses the selective exclusion of semantically inconsequential masked tokens from the weight aggregation process pertaining to visible tokens with (\ref{eq:spa_loss}), and at the same time, it enforces data singularity constraints (\ref{eq:entropy_maximization}) and (\ref{eq:rank_loss}) based on the depth of the layer to enhance the model's capability to accurately identify regions necessitating semantic restoration.
    }
    \label{fig:method}
    \vspace{-0.6cm}
\end{figure}

\begin{equation}
    \mathcal{L}_{spa}(f(I;\Theta)) = -\sum_{i \notin \delta_{M}}\sum_{j}(p_{i,j} \log{p_{i,j}}) 
    \label{eq:spa_loss}
\end{equation}

\noindent where $p$ is an element of affinity matrix $\psi(X X^{\top})$ that satisfies $0 < p_{i,j} < 1$ and $\sum_{j}{p_{i,j}} = 1$. 

Minimizing the $\mathcal{L}_{spa}$ loss on a row-wise basis ensures the preferential allocation of maximum weight amongst the interaction of visible and visible tokens rather than the interaction of visible and masked tokens, 
which facilitates the exclusion of semantically inconsequential masked tokens from the weight aggregation process of visible tokens. This assertion can be substantiated through the following proof. For the sake of simplicity, we omit the row index $i$ of $p$ as we only consider a single-row probability of the affinity matrix in the proof.






\begin{proof}

Let $\mathbf{U} = \left\{ x \,|\, x \in \mathbf{A} \right\} \cup \left\{ y \,|\, y \in \mathbf{B} \right\}$ be a single row of the affinity map where the $x$-axis and $y$-axis are both arranged in the order of masked token $X_M$ and visible image token $X_V$. Without loss of generality, we assume that $\mathbf{A} = \left\{ p_1, p_2, \ldots, p_{n-1}, p_n \right\}$ and $\mathbf{B} = \left\{ p_{n+1}, p_{n+2}, \ldots, p_{N-1}, p_N \right\}$ refers to the weights between the visible-masked tokens and visible-visible tokens, respectively. Here, $n=|\delta_M|$ and $N$ indicate the number of masked tokens and the number of all tokens, respectively. The entropy of $U$ is expressed as $\mathbb{E}_\mathbf{U} = -\sum_{l=1}^N(p_l \log{p_l})$. It can be decomposed as $\mathbb{E}_\mathbf{U} = \mathbb{E}_\mathbf{A} + \mathbb{E}_\mathbf{B}$, 
where $\mathbb{E}_\mathbf{A} = -\sum_{l=1}^n(p_l \log{p_l})$ and $\mathbb{E}_\mathbf{B} = -\sum_{l=n+1}^N(p_l \log{p_l})$.


\noindent
The following conditions apply:\\
\noindent \textbf{Condition 1.} $p_1 = p_2 = \ldots = p_{n-1} = p_n$\\
\noindent \textbf{Condition 2.} $\forall p \in \mathbf{A} : 0 < p < \frac{1}{n}$\\
\noindent \textbf{Condition 3.} if $\max{p_i} \in \mathbf{A} : \max{p_i} \approx \frac{1}{n}$,\\
\indent \quad \quad \quad \quad if $\max{p_i} \in \mathbf{B} : \max{p_i} \approx 1$,\\
\indent \quad \quad \quad \quad $\min{p_j} = \varepsilon \ (i\ne j, \ 1\le i \le N, \ 1\le j \le N)$\\

Condition 1 adheres to the property of substitutional consistency, which is the second in the properties in Section~\ref{sec:intro}, and the term $\epsilon$ represents an infinitesimally small value. When $\max{p_i} \in \mathbf{B}$, the value of $\mathbf{\mathbb{E}_\mathbf{U}}$ is lower than when $\max{p_i} \in \mathbf{A}$. The corresponding proposition has been validated through inductive reasoning. A simplified version of the case of $n=k$ is provided in the manuscript, and the complete proof is detailed in the supplementary material.\\

\noindent In case $n = k$\\
 $\mathbf{A} = \left\{ p_1, p_2, \ldots, p_{k-1}, p_k\right\}$,
 $\mathbf{B} = \left\{p_{k+1}, p_{k+2}, \ldots, p_t, \ldots, p_{N-1}, p_N \right\}$

\item By \textbf{\textit{Condition 2.}} $\forall p \in \mathbf{A}: 0 < p < \frac{1}{k}$\\

\vspace{-0.3cm}
\item \lowercase\expandafter{\romannumeral1}) \textbf{Case 1} \\
If $\max{p_i} \in \mathbf{A}$ \\
By \textbf{\textit{Condition 1.}} and \textbf{\textit{Condition 2.}} $\forall p \in \mathbf{A} : p \approx  \frac{1}{k}$ \\
$p_{k+1} = p_{k+2} = \cdots = p_{N-1} = p_N = \varepsilon$ 

\vspace{-0.4cm}
\begin{align}
\label{eqn:my_equation}
    \begin{split}
        \mathbb{E}_\mathbf{U} &= \mathbb{E}_\mathbf{A} + \mathbb{E}_\mathbf{B} 
        = \sum_{l=1}^k(- p_l \log{p_l}) + \sum_{l=k+1}^N(- p_l \log{p_l}) \\
        &\to \log{k}
    \end{split}
\end{align}
\vspace{-0.3cm}
\item \lowercase\expandafter{\romannumeral2}) \textbf{Case 2} \\
If $\max{p_i} \in \mathbf{B}$ \\
$p_t = \max{p_i} \ (n+1 \le t \le N)$, by \textbf{\textit{Condition 3.}} $p_t \approx 1$\\
By \textbf{\textit{Condition 3.}} $\forall p \in \mathbf{A}: p < p_t$ \\
$p_t \approx 1 - \sum_{l=1}^k p_l$ \\
$\therefore p_1 = p_2 = \cdots = p_{k-1} = p_k = \varepsilon$, \quad $p_{k+1} = p_{k+2} =\cdots = p_{t-1} = p_{t+1} = \cdots = p_{N-1} = p_N = \varepsilon$ 

\vspace{-0.3cm}
\begin{allowdisplaybreaks}
\begin{align}
\label{eqn:my_equation}
    \begin{split}
        \mathbb{E}_\mathbf{U} &= \mathbb{E}_\mathbf{A} + \mathbb{E}_\mathbf{B}
        = \sum_{l=1}^k(- p_l \log{p_l}) +
        \left\{- p_t \log{p_t} + \sum_{l=k+1, l \ne t}^N(- p_l \log{p_l})\right\} \\
        &\to 0
    \end{split}
\end{align}
\end{allowdisplaybreaks}

\vspace{-0.3cm}
Always \quad $\mathbb{E}_{\mathbf{U}_{case1}} > \mathbb{E}_{\mathbf{U}_{case2}}$ 
\end{proof}

The above proof explicates the underlying operational principle of (\ref{eq:spa_loss}), elucidating that achieving minimum entropy per row is contingent upon the consistent assignment of maximum weight exclusively to interactions between visible tokens. This strategic approach plays a pivotal role in effectively preventing the influence of masked token values on the representation learning of visible tokens, thereby upholding the integrity of the learning algorithm.

Furthermore, as investigated in Section~\ref{sec:analysis}, the parameter pertaining to the masked token of the initial embedding is trained to exhibit a diminutive correlation with the visible token to fulfill the property of data singularity. In existing methods, this property can solely be achieved by means of the model's convergence, which is secured through a long pre-training procedure. From the perspective of the distinctiveness of the masked token in the initial embedding, we propose to explicitly augment the heterogeneity from the visible token rather than solely relying on model convergence. By distinctly differentiating the masked token from the visible token, the network gains the capability to accurately identify regions necessitating semantic restoration, thereby paving the way for a more efficient learning process. The optimization for the initial masked token embedding can be formulated as:


\begin{equation}
    \mathcal{L}_{e}(f(I;\Theta))=\frac{1}{H^{0} + \epsilon},
    \label{eq:entropy_maximization}
\end{equation}
where $H^{0}$ denotes heterogeneity defined in (\ref{eq:heterogeneity_measure}) of the initial embedding stage before passing through the attention layers and $\epsilon$ is a small value to prevent zero division. 
Eq. (\ref{eq:entropy_maximization}) augments the distinctiveness of masked tokens by maximizing the heterogeneity of masked tokens over visible tokens. 

On the other hand, masked tokens in the subsequent layers tend to exhibit a notable correlation towards the visible tokens as they are gradually reconstructed through interaction with neighboring tokens in the attention layers. In light of this, with regard to the distinctiveness among tokens, we impose a constraint on the subsequent layers to have progressively lower heterogeneity. Considering the common direction of both aspects, we pursue a gradual reduction of heterogeneity. To this end, we intuitively employ the form of a ranking loss, strategically applied to the subsequent layers:
\begin{equation}
    \mathcal{L}_{r}(f(I;\Theta)) = \sum_{l=1}^{L} \log (1+\exp(H^{l-1} - H^{l})).
    \label{eq:rank_loss}
\end{equation}

By means of Eq. (\ref{eq:rank_loss}), we enforce a constraint upon the masked tokens in the subsequent layers, forcing them to exhibit diminished distinguishability from the visible tokens. Concurrently, we grant masked tokens the capability to exert an enhanced influence over the feature-learning process of the visible tokens. This intricate interplay of masked tokens across entire layers strikes a balance, promoting the convergence of token representations while employing only essential information within the learning framework.


\section{Experiments}
In this section, we assess the efficacy of the proposed MTO approach through a series of pre-training and fine-tuning experiments. As MTO is an adaptable and plug-and-play method for any Masked Image Modeling (MIM)-based approach that utilizes masked tokens, we apply it to multiple baseline approaches~\cite{xie2022simmim, he2022masked, dong2022bootstrapped, yi2022masked} to evaluate the effectiveness. Please refer to the Supplementary material for more experimental results, detailed analysis, and ablation studies.

\subsection{Metric for Efficient Pre-training}
The main objective of MTO is to reduce the substantial pre-training time of Transformer-based architecture, that is to say, accelerating the convergence speed. The area under the curve can be one of the metrics that quantify the rate of convergence because the faster the network converges and the better the network performance, the higher the value. Thus, to quantify the relative performance improvement over the baseline approaches, we propose the RAUC measure, denoting the relative area under the curve, as follows: 


%

\begin{equation}
    {RAUC}(S1, S2; E1, E2) = \frac{\int_{E1}^{E2} (S2(E) - S1(E1)) dE}{ \int_{E1}^{E2} (S1(E) - S1(E1)) dE}
    \label{eq:rauc}
\end{equation}

Here, $E1$ and $E2$ represent the number of epochs, and $S(E)$ is set to the performance of the target method $S$ at specific epoch $E$. 
This measure serves as a quantitative indicator, precisely delineating the extent of relative performance improvement across a specified range of epochs.

\subsection{Baseline Models}
\noindent{\bf{SimMIM}}~\cite{xie2022simmim} models masked image reconstruction as a pretext task for self-supervised pre-training of Transformer architecture. In SimMIM, masked tokens are semantically encoded with visible image tokens in the Transformer encoder and are reconstructed with shallow MLP layers. We pre-train the ViT-B following the same hyper-parameters as \cite{xie2022simmim} and our losses are additionally adopted to the Transformer encoder.

\noindent{\bf{MAE}}~\cite{he2022masked}. Different from SimMIM~\cite{xie2022simmim}, only visible image tokens pass through the Transformer encoder for efficient pre-training. Thus, masked tokens are concatenated with encoded visible tokens and pass through the Transformer decoder, separating the semantic encoding task from the pretext prediction task. In MAE, We pre-train the ViT-B and ViT-L following the same hyper-parameters as \cite{he2022masked} and losses are adopted to the Transformer decoder.

\noindent{\bf{BootMAE}}~\cite{dong2022bootstrapped} introduced bootstrapped MAE that combines a bootstrapped feature prediction task into the original MAE. BootMAE learns separate decoders for pixel regression and feature prediction with the same masking strategy as MAE. Thus, we adopted our masked token optimization strategy for both pixel regression and feature prediction Transformer decoders.

\noindent{\bf{ConMIM}}~\cite{yi2022masked} taps into the significant potential of contrastive learning within denoising auto-encoding frameworks. It focuses on generating straightforward intra-image inter-patch contrastive constraints as the primary learning goals for predicting masked patches. This approach eliminates the need for additional training stages often required in customizing image tokenizers.

\begin{figure*}[t!]
    \centering
    \begin{subfigure}{0.31\linewidth}
        \includegraphics[width=\linewidth]{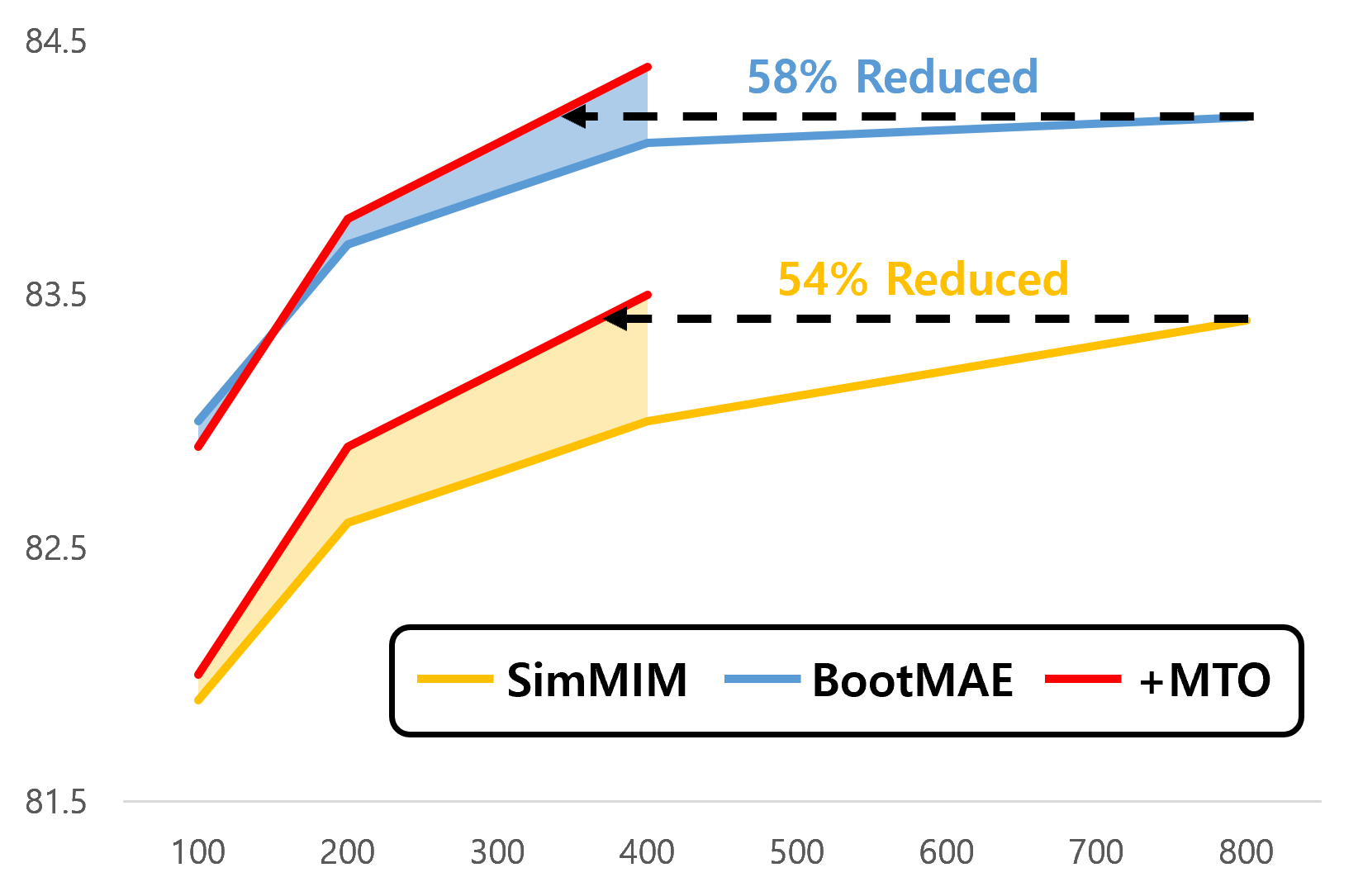}
    \end{subfigure}
    \hfill
    \begin{subfigure}{0.31\linewidth}
        \includegraphics[width=\linewidth]{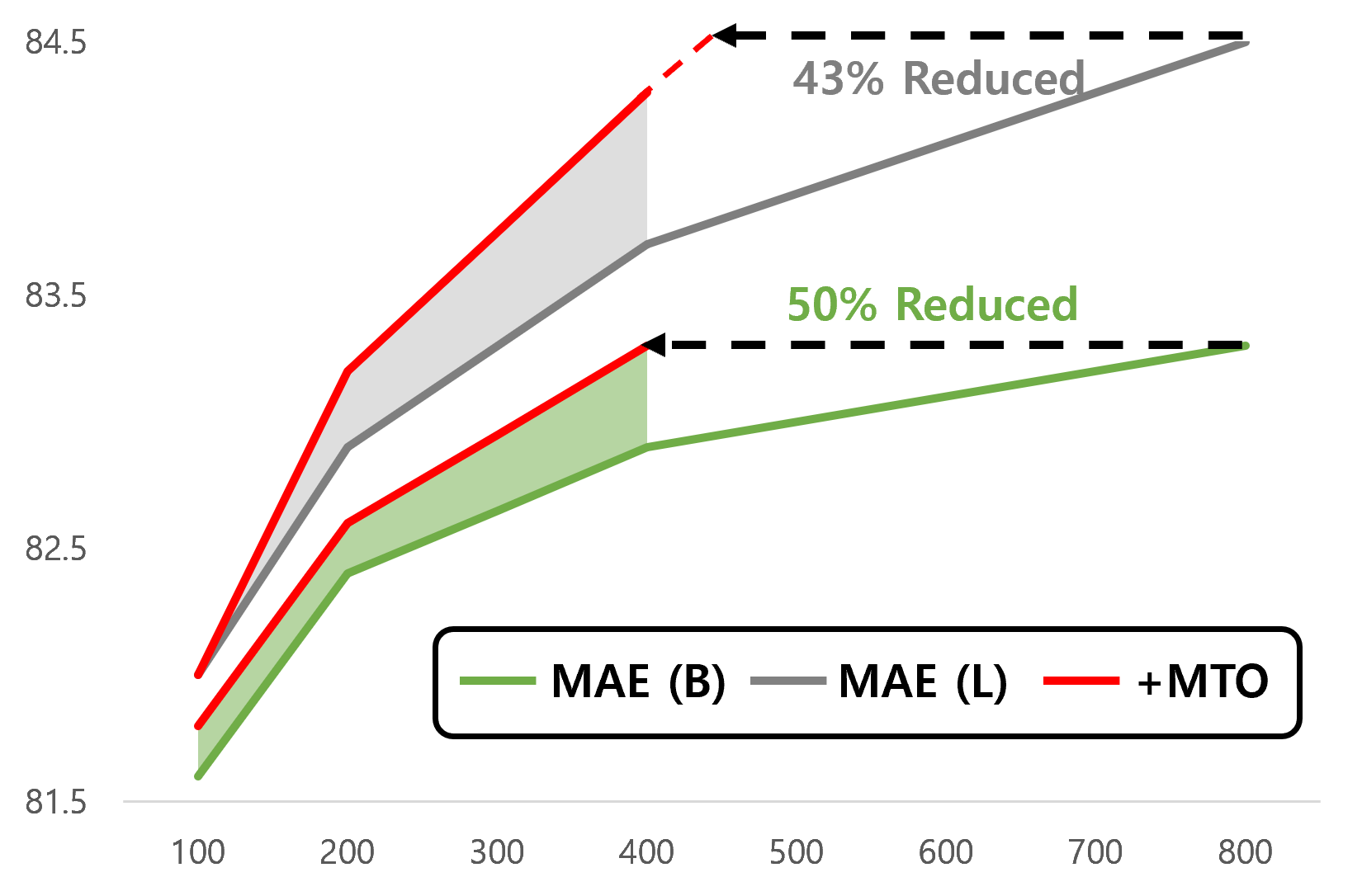}
    \end{subfigure}
    \hfill
    \begin{subfigure}{0.31\linewidth}
        \includegraphics[width=\linewidth]{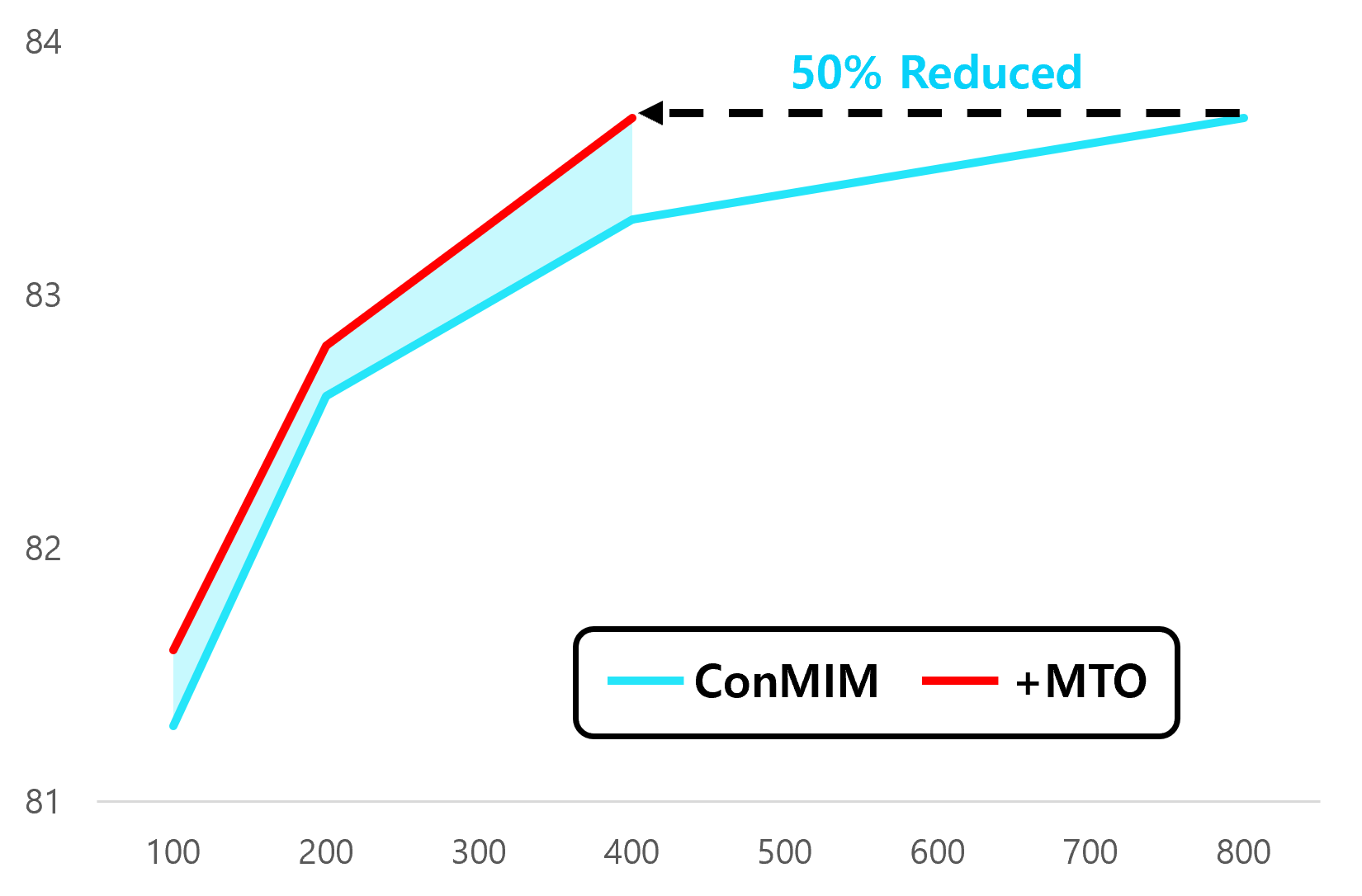}
    \end{subfigure}
    \centering
    \caption{The comprehensive performance results of applying MTO to various baselines~\cite{xie2022simmim, he2022masked, dong2022bootstrapped, yi2022masked}. MTO achieves a substantial improvement in the efficiency of pre-training by attaining the standard performance within approximately 400 epochs across all baseline methods in common. This signifies that remarkable enhancement in efficiency is achievable across any MIM method through the application of MTO, rendering it a viable option for masked tokens.
    }
    \label{fig:main_result}
    \vspace{-0.3cm}
\end{figure*}

\begin{table*}[t]
    \centering
    \addtolength{\tabcolsep}{8pt}
    \renewcommand{\arraystretch}{0.9}
    \begin{tabular}{cc|ccc}
        \toprule
        \multicolumn{4}{c}{${RAUC}(S1, S2; E1, E2)$}   \\ \midrule
        \multirow{2}{*}{$S1$}   &   \multirow{2}{*}{$S2$}    &   \multicolumn{2}{c}{$(E1, E2)$}  \\ 
        {}  &   {}  &   (100, 400) &   (200, 400) \\
        \midrule
        SimMIM~\cite{xie2022simmim}  &   SimMIM + MTO    &  1.47	& 1.44 \\  
        BootMAE~\cite{dong2022bootstrapped} &   BootMAE + MTO    & 1.19	& 1.22 \\
        
        MAE~\cite{he2022masked}  &   MAE + MTO    & 1.32	& 1.29 \\
        ConMIM~\cite{yi2022masked}  &   ConMIM + MTO    & 1.21	& 1.17 \\
        \midrule
    \end{tabular}
    \caption{We report the evaluation of the proposed relative area under the curve ($RAUC$) measure over the baseline approaches~\cite{xie2022simmim, he2022masked, dong2022bootstrapped, yi2022masked}. The same backbone network (ViT-B) is used for pre-training.}
    \label{tab:RAUC_result}
    \vspace{-0.5cm}
\end{table*}

\begin{table*}[t!]
\centering
\addtolength{\tabcolsep}{12pt}
\renewcommand{\arraystretch}{0.9}
\begin{tabular}{ccc|c}
\hline
\multicolumn{3}{c|}{Proposed optimization}                  & \multirow{2}{*}{Top-1 Acc(\%)} \\ \cline{1-3}
$\mathcal{L}_{spa}$ & $\mathcal{L}_{e}$ & $\mathcal{L}_{r}$ &                                \\ \hline
$\checkmark$        &                   &                   & 83.2                           \\
                    & $\checkmark$      &                   & 83.3                           \\
                    &                   & $\checkmark$      & 82.8                           \\
$\checkmark$        & $\checkmark$      &                   & 83.4                           \\
$\checkmark$        &                   & $\checkmark$      & 83.0                           \\
                    & $\checkmark$      & $\checkmark$      & 83.4                           \\
$\checkmark$        & $\checkmark$      & $\checkmark$      & 83.5                           \\ \hline
\end{tabular}
\vspace{0.2cm}
\caption{We present the detailed ablation study conducted for the importance of the proposed objectives.
}
\label{tab:loss_balancing}
\vspace{-0.9cm}
\end{table*}

\subsection{Performance Comparisons}\label{sec:performance_comparison}
We report the results of the proposed method trained on the recent baselines~\cite{xie2022simmim, he2022masked, dong2022bootstrapped, yi2022masked} in Figure~\ref{fig:main_result} and Table~\ref{tab:RAUC_result}.
Following the same settings for each baseline method, we pre-trained and fine-tuned on ImageNet-1K classification dataset for main evaluation. \textcolor{hs}{Note that, training recent approaches require a huge hardware specification, making it increasingly difficult to reproduce the reported results. As all experiments were conducted in our hardware configuration ($8 \times$ RTX 3090) for fair comparison, the results we reported may differ from those of their manuscript. 
For the fair experimental schedule, all the 400 epoch performances were equally measured in intermediate stages in the training towards 800 epochs.}

Figure~\ref{fig:main_result} presents a main comparison of the top-1 accuracy between the baseline models of SimMIM~\cite{xie2022simmim}, MAE~\cite{he2022masked}, BootMAE~\cite{dong2022bootstrapped}, ConMIM~\cite{yi2022masked} and our method with the MTO applied. We conducted experiments on ViT-B~\cite{dosovitskiy2020image} and ViT-L as a backbone attention network.

Through the application of MTO, the convergence process was significantly accelerated in all baseline methods, with the baseline's standard performance being achieved within the range of 300 to 500 epochs. Specifically, in the case of the SimMIM~\cite{xie2022simmim} and BootMAE~\cite{dong2022bootstrapped} baseline, the application of MTO remarkably expedited performance, achieving in merely 300 to 400 epochs what typically requires 800 epochs, thereby realizing an impressive pre-training reduction rate of 54\% and 58\%. For both MAE~\cite{he2022masked} and ConMIM~\cite{yi2022masked}, the performances of the existing 800-epoch baseline were attained remarkably within just 400 epochs with the application of MTO. Our approach facilitated a substantial pre-training reduction rate of 50\% for both methods. Moreover, even in the MAE~\cite{he2022masked} baseline utilizing the data-hungry model ViT-L, the introduction of MTO yielded impressive outcomes, manifesting in a pre-training epoch reduction rate of 43\%. 

The deployment of the proposed methodology across a range of baselines unequivocally validated the improvement of pre-training efficiency attributed to the implementation of MTO. This advancement stemmed from the effective training of masked tokens, a theme emphasized consistently throughout the manuscript, leading to a marked enhancement in the overall efficacy of pretext prediction and the refinement of representation learning.

\textbf{$RAUC$}: Table~\ref{tab:RAUC_result} reports the evaluation using the proposed relative area under the curve ($RAUC$) measure over the baseline approaches~\cite{xie2022simmim, he2022masked, dong2022bootstrapped, yi2022masked}. This introduced metric effectively highlights the relative performance enhancements, offering a clear and immediate understanding upon initial observation. Across the range from 200 to 400 epochs, the relative performance improvement rates for SimMIM~\cite{xie2022simmim}, BootMAE~\cite{dong2022bootstrapped}, MAE~\cite{he2022masked}, and ConMIM~\cite{yi2022masked} are 44\%, 22\%, 29\% and 17\%, respectively. Besides, in the overall range from 100 to 400 epochs, the relative performance increase rate was amplified further, showing rates of 47\%, 19\%, 32\%, and 21\% for each baseline. The result accentuates the diverse degrees of enhancement achieved by applying MTO to each method. Moreover, it confirms MTO's efficacy in reducing the pre-training epochs across all baselines, encompassing the full spectrum of the training procedure.

\subsection{Ablation on Objectives}

In our approach, we introduce three novel objectives, denoted as `$\mathcal{L}_{spa}$', `$\mathcal{L}_{e}$' and `$\mathcal{L}_{r}$' specifically designed for masked token optimization. Table~\ref{tab:loss_balancing} presents the detailed ablation study conducted for the importance of these objectives. All experiments report the ImageNet-1K classification accuracy on 400 epochs of SimMIM~\cite{xie2022simmim} using ViT-B as a backbone architecture.

Broadly speaking, each objective contributed significantly to enhancing the performance and expediting the convergence process. Upon delving into the specifics, it becomes apparent that $\mathcal{L}_{r}$, when utilized in isolation, emerged as a factor contributing to the destabilization of performance outcomes. This phenomenon arises due to the fact that ranking loss merely modulates the magnitude of heterogeneity for each layer sequentially. Such a singular approach carries the risk of systemic collapse, potentially leading to scenarios where all heterogeneity converges to zero, thus undermining the model's structural integrity. Nevertheless, the simultaneous application of $\mathcal{L}_{r}$ with $\mathcal{L}_{e}$ leads to a harmonized effect. The entropy maximization impact of $\mathcal{L}_{e}$ at the first layer acts as a regulatory mechanism, effectively elevating the overall performance to a notable 83.4. Moreover, both $\mathcal{L}_{spa}$ and $\mathcal{L}_{e}$ demonstrate significant importance within the overall methodology. The sole application of each led to favourable enhancements in overall performance, elevating it to 83.2 and 83.3, respectively, underscoring their individual efficacy in the process. Consequently, the synergistic integration of all three optimizations emerged as a requisite for achieving the zenith of convergence acceleration and performance enhancement, highlighting the necessity of their collective implementation for optimal results.

\section{Related Work}

\subsection{Masked Language Modeling}
Masked Language Modeling (MLM)~\cite{devlin2018bert, liu2019roberta, clark2020electra, bao2020unilmv2, zaken2021bitfit, ghazvininejad2019mask, song2020mpnet, song2019mass, raffel2020exploring, conneau2019cross} predicts removed tokens based on remained ones to inject the ability of learning semantic representation of a corpus to the network.
While MLM has brought rapid advances in natural language processing (NLP) and have been shown to scale and generalize well on downstream tasks~\cite{brown2020language}, 
the problem of prelonged convergence time and immense computation of naive MLM still remained. Amid these challenges, more efficient self-supervised pre-training approaches~\cite{lanalbert, chen2020earlybert, wettig2022should, lee2022efficient, liao2022mask} have been proposed.
ALBERT~\cite{lanalbert} proposes two parameter reduction techniques,   factorized embedding parameterization and cross-layer parameter sharing, for memory efficiency and shortening the training time. 
Based on the Lottery Ticket Hypothesis (LTH), EaryBERT~\cite{chen2020earlybert} prunes the network for efficient pre-training and fine-tuning. CCM~\cite{lee2022efficient} designs a curriculum masking framework that gradually masks similar tokens of similar concepts in an easy-to-difficult order. Similar to MAE~\cite{he2022masked}, 3ML~\cite{liao2022mask} learns the encoder by separating the mask tokens from the sequence and conducts reconstruction only through the decoder.


        

\subsection{Masked Image Modeling}
Masked Image Modeling (MIM)~\cite{chen2020generative, v1997determination, bao2021beit, zhou2021ibot, he2022masked, xie2022simmim, huang2022green, cao2022understand, dong2022bootstrapped, zhang2023hivit, liu2022mixmim, pan2022towards, zhang2022mask, peng2022beit, peng2022unified, hou2022milan, xue2023stare} is a relatively new technique that has gained popularity in the field of computer vision and machine learning in recent years. The basic idea behind MIM is to predict missing or occluded parts of an image using a neural network trained on partially masked images. Inspired by NLP, iGPT~\cite{chen2020generative} and iBERT~\cite{v1997determination} have attempted to transfer the pretext task of masked prediction from language to image data, but these have caught less attention due to their inferior performance to other approaches. Different from iBERT which directly reconstructs the masked patches, BEiT~\cite{bao2021beit} uses a two-stage approach that requires a pre-trained discrete variational autoencoder (dVAE) to generate discretized target visual tokens. In contrast, MAE~\cite{he2022masked} and SimMIM~\cite{xie2022simmim} are end-to-end training methods of masked autoencoders. MAE predicts masked patches directly from unmasked ones with a high masking ratio of 75\%. SimMIM has a similar structure to MAE but with a larger patch size and multiple masking strategies.

Despite the impressive performance, masked autoencoder approaches require a large amount of computation with large-scale training datasets. Researchers have explored using hierarchical Vision Transformers (ViTs) to improve the efficiency of pre-training models for masked image modeling by enabling the ViTs to discard masked patches and only operate on visible ones. GreenMIM~\cite{huang2022green} introduced group window attention, while HiViT~\cite{zhang2023hivit} and MixMAE~\cite{liu2022mixmim} enable masking in hierarchical ViT. In contrast to prior methodologies, the proposed method considers the inherent properties of the tokens employed by MIM as a fundamental approach to effective pre-training.


\section{Conclusion}
This work delves into the properties of masked tokens, examines their heterogeneity with visible tokens, and proposes a novel approach termed masked token optimization (MTO). \textcolor{hs}{MTO boosts both pretext prediction and semantic encoding by emphasizing data singularity of the masked token, achieving a considerable improvement on pre-training efficiency.
Also, our method can be applied to any method in a plug-and-play manner thanks to a simple approach that only adds loss functions.}

\noindent{\bf{Limitations}}
Within the scope of this manuscript, the triad of properties attributed to masked tokens, are not to be deemed immutable, but rather dynamic concepts subject to evolution. With the continual progression in the realm of Masked Image Modeling, it becomes imperative that these attributes undergo constant updates and refinement. Embracing this process of perpetual revision and advancement is vital to remain abreast of the ever-evolving landscape of research in this specialized field.

\clearpage
\clearpage
\clearpage

%
%
\bibliographystyle{splncs04}
\bibliography{main}
\end{document}